\documentclass[letterpaper]{article} 
\usepackage{times} 
\usepackage{helvet}
\usepackage{courier}
\usepackage[hyphens]{url} 
\usepackage{graphicx}

\title{Beyond the Grounding Bottleneck:\\ Datalog Techniques for
  Inference in Probabilistic Logic Programs\\ (Technical Report)}

\author{Efthymia Tsamoura, Samsung AI Research\\  V\'ictor
  Guti\'errez-Basulto, Cardiff University\\ Angelika Kimmig, Cardiff University }

\usepackage{algorithm}
\usepackage[noend]{algpseudocode}
\usepackage{colonequals} 
\usepackage{amsmath, amsfonts, amssymb}
\usepackage{paralist}
\usepackage{soul}
\usepackage{color}
\usepackage[square]{natbib}

\newcommand{\Tcp}{\mathit{Tc}_{\mathcal{P}}}
\newcommand{\ok}[1]{#1}
\newcommand{\vp}{\ensuremath{\mathsf{vProbLog}}}
\newcommand{\Pp}{\mathcal{P}}
\newcommand{\Rp}{\mathcal{R}}
\newcommand{\Fp}{\mathcal{F}}
\newcommand{\DNF}{\lambda}
\DeclareMathOperator{\WMC}{WMC}
\newcommand{\Tp}{T_{\mathcal{P}}}
\newcommand{\Ap}{\mathsf{HB}(\Pp)}
\newcommand{\Ht}{\mathcal{I}}
\newcommand{\Hi}{\mathit{I}}
\newtheorem{definition}{Definition}
\newtheorem{theorem}{Theorem}
\newcommand*{\defeq}{\mathrel{\mathop:}=}
\newcommand{\magic}[1]{\mathsf{magic}(#1)}
\newcommand{\Scp}{\mathit{Sc}_{\mathcal{P}}}
\newcommand{\Mp}{\mathcal{M}}
\newcommand{\Tcm}{\mathit{Tc}_{\mathcal{M}}}
\newcommand{\Mcp}{\mathit{Mc}_{\mathcal{P}}}
\newcommand{\MAp}{\mathsf{HB}(\Mp)}
\newcommand{\vpplain}{\ensuremath{\mathsf{vProbLog^{plain}}}}
\newcommand{\vpopt}{\ensuremath{\mathsf{vProbLog^{opt}}}}
\newcommand{\problog}{\ensuremath{\mathsf{ProbLog2}}}
\newcommand{\lubm}{\textbf{LUBM}}
\newcommand{\lubmS}{\textbf{LUBM001}}
\newcommand{\lubmM}{\textbf{LUBM010}}
\newcommand{\lubmL}{\textbf{LUBM100}}
\newcommand{\webkb}{\textbf{WebKB}}
\newcommand{\smokers}{\textbf{Smokers}}
\newcommand{\genes}{\textbf{Genes}}
\newcommand{\ad}[1]{\mathsf{#1}}
\newcommand{\mgc}[2]{m_{#1}^{#2}}
\newcommand{\predQ}{\mathtt{q}}
\newcommand{\vars}[1]{\mathsf{vars}(#1)}
\newcommand{\RMp}{\Rp_{\Mp}}
\newtheorem{lemma}{Lemma}

\begin{document}

\maketitle

\begin{abstract}
State-of-the-art inference approaches in probabilistic logic
programming typically start by computing the relevant ground program
with respect to the queries of interest, and then use this program for
probabilistic inference using knowledge compilation and weighted model
counting. We propose an alternative approach that uses
efficient Datalog techniques to integrate knowledge compilation
with forward reasoning with a non-ground program. This effectively
eliminates the grounding bottleneck that so far has prohibited the
application of probabilistic logic programming in query answering
scenarios over knowledge graphs, while also providing fast approximations on
classical benchmarks in the field. 
\end{abstract}

\section{Introduction}\label{sec:intro}
The significant interest in combining logic and probability for
reasoning in uncertain, relational domains has led to a multitude of
formalisms, including the family of probabilistic logic programming
(PLP) languages based on the distribution
semantics~\citep{sato:iclp95} with languages and systems such as
PRISM~\citep{sato:iclp95}, ICL~\citep{poole:pilp08},
ProbLog~\citep{deraedt:ijcai07,fierens:tplp15} and
PITA~\citep{riguzzi:tplp11}. 
State-of-the-art inference for PLP uses a reduction to weighted model
counting (WMC)~\citep{chavira:ai08}, where the dependency structure of
the logic program and the queries is first transformed into a
propositional formula in a suitable format that supports efficient
WMC. While the details of this transformation differ across
approaches, a key part of it is determining the relevant ground
program with respect to the queries of interest, i.e., all groundings
of rules that contribute to some derivation of a query. 
This grounding step has received little attention, as its cost is dominated by the cost of
constructing the propositional formula in 
typical PLP
benchmarks that operate on biological, social or hyperlink networks,
where formulas are complex. 
However, it has been observed that the grounding
step is the bottleneck that often makes it impossible to apply PLP
inference in the context of ontology-based data access over
probabilistic data (pOBDA) \citep{schoenfisch:ijar17,vBDJ-CIKM19}, where
determining the relevant grounding explores a large
search space, but only small parts of this space contribute to the formulas.

We address this bottleneck, building upon the $\Tcp$ operator
  \citep{vlasselaer:ijcai15}, which integrates formula construction
  into forward reasoning for ground programs and is
state-of-the-art for highly cyclic PLP programs. Our key contribution
is a program transformation
approach that allows us to implement forward inference using an efficient Datalog engine that
directly operates on non-ground functor-free programs.
We focus on programs without
negation for simplicity, though the $\Tcp$ operator has been studied
for general probabilistic logic programs
\citep{bogaerts:tplp15,riguzzi:ijar16} as well; the extension 
to stratified negation following \citep{vlasselaer:ijar16} is
straightforward. We further build upon two well-known techniques
from the Datalog community, namely semi-naive evaluation~\citep{AbiteboulHV95}, which avoids
recomputing the same consequences repeatedly during forward reasoning, and the magic
sets transformation \citep{bancilhon:pods86,beeri:jlp91}, which makes forward reasoning query driven. We
adapt \ok{and extend} both techniques to \ok{incorporate the formula
  construction performed by} the $\Tcp$ operator and implement our approach
using VLog \citep{urbani:aaai16,CDGJKU2019}. Our experimental evaluation demonstrates that the
resulting \vp\ system enables PLP inference in the
pOBDA setting, answering each of the 14 standard queries of the LUBM
benchmark \citep{lubm} over a probabilistic database of 19K facts in a
few minutes at most, while most of these are infeasible
for the existing ProbLog implementation of $\Tcp$. Furthermore, for
ten of the queries, \vp\ computes exact answers over 1M facts in
seconds. At the same time, on three standard PLP benchmarks~\citep{fierens:tplp15,renkens:aaai14,vlasselaer:ijar16} where the
bottleneck is formula construction, \vp\ achieves comparable
approximations to the existing implementation in less time. 

We provide details on proofs as well as additional background 
in the Appendix.

\section{Background}\label{sec:bg}
We provide some basics on probabilistic logic programming. We use standard notions of  propositional logic and logic programming, cf.\ Appendix.

We focus on the  probabilistic logic programming language ProbLog
\citep{deraedt:ijcai07,fierens:tplp15}, and \ok{consider only function-free
logic programs.}

A  \emph{rule} (or \emph{definite clause}) is a universally quantified expression of the form $\mathtt{h \colonminus b_1, ... , b_n}$ where
$\mathtt{h}$ and the $\mathtt{b_i}$ are atoms and the comma denotes conjunction.
A \emph{logic program} (or \emph{program} for short) is a finite set of rules.
A \emph{ProbLog program} $\Pp$ is a triple $(\Rp,\Fp, \pi)$,
  where  $\Rp$ is a \emph{program}, $\Fp$ is a finite set of ground facts\footnote{Note
  that the semantics is well-defined for countable 
  $\Fp$, but assume (as usual in exact inference) that the finite
  support condition holds, which allows us to restrict to finite $\Fp$
  for simplicity.} and
  $\pi: \Fp \rightarrow [0,1]$ a function that labels facts with
  probabilities, which is often written using annotated facts $p::f$
  where $p=\pi(f)$. 
 Without loss of generality, we
  restrict $\Rp$ to non-fact rules and include `crisp' logical facts $f$  in
  $\Fp$ by setting $\pi(f)=1$. We also refer to a ProbLog
  program as probabilistic program. As common in probabilistic logic programming (PLP), we assume that the sets of predicates defined by  facts in $\Fp$
and rules in $\Rp$, respectively, are disjoint.
A ProbLog program specifies a probability distribution over its
Herbrand interpretations, also called \emph{possible worlds}. Every fact $f\in \Fp$ independently takes values $\mathtt{true}$ with probability $\pi(f)$
or $\mathtt{false}$ with probability $1-\pi(f)$.

For the rest of the section we fix a probabilistic program $\Pp=(\Rp,\Fp, \pi)$.
A \emph{total choice} $C\subseteq\Fp$ 
 assigns a
truth value to every (ground)  fact, and the
corresponding logic program $C\cup\Rp$ has a unique least Herbrand
model; the
probability of this model is that of $C$. Interpretations that do not correspond to any total choice
have probability zero. The \emph{probability of a query $q$} is then the sum
over all total choices whose program entails $q$:
\begin{align}
\label{eq:p_suc}
\Pr(q) &\colonequals \sum\limits_{C\subseteq \Fp:   C\cup \Rp\models q}\, \prod_{f \in C} \pi(f) \cdot\prod_{f\in \Fp\setminus C}(1-\pi(f))\;.
\end{align}

As enumerating all total choices entailing the query is infeasible,
state-of-the-art ProbLog inference reduces the problem to that of weighted model counting.
For a formula $\DNF$ over propositional variables $V$ and a weight function $w(\cdot)$ assigning a real number to every literal for an atom in $V$, the weighted model count is defined as
\begin{align}
\label{eq:wmc}
\WMC(\DNF) &\colonequals \sum\limits_{I\subseteq V :
    I\models \DNF}\, \prod_{a\in I}w(a)\cdot\prod_{a\in V\setminus
  I}w(\neg a)\;.
\end{align}
The reduction assigns $w(f)=\pi(f)$ and $w(\neg f)=1-\pi(f)$ for
 facts $f \in \Fp$, and $w(a)=w(\neg a) = 1$ for other atoms. For
a query $q$, it constructs a formula $\DNF$ such that for every total choice
$C\subseteq \Fp$, $C \cup \{\DNF\} \models q$ if and only if  $C\cup\Rp\models
q$. While $\DNF$ may use variables besides the 
\ok{facts}, e.g., variables corresponding to the atoms in the program, their values have to be
uniquely defined for each total choice.

\smallskip
\ok{In what follows we present (and adapt)  notions and results by  \citeauthor{vlasselaer:ijar16}~(\citeyear{vlasselaer:ijar16}).}

\ok{We start by noting that} one way to (abstractly) specify such a $\DNF$ is to take a disjunction over the conjunctions of facts in all total choices that entail the \ok{query} of interest\ok{:
\begin{align*}\label{eq:target}
 \bigvee\limits_{C\subseteq \Fp:   C\cup
         \Rp\models q}\, \bigwedge_{f\in
         C}f
\end{align*}}
\ok{We next extend the immediate consequence operator $\Tp$  for
  classic logic programs to construct parameterized interpretations associating
a propositional formula with every atom.  }

\ok{Recall   that  the $\Tp$ operator is used to  derive new knowledge starting from the facts. Let $\Pp$ be a logic program.
For a Herbrand interpretation $\Hi$, the $\Tp$ operator returns
\[
\Tp(\Hi) = \{ \mathtt{h} \mid \mathtt{h \colonminus b_1,\ldots,b_n} \in \Pp \mbox{ and } \{ \mathtt{b_1}, \ldots , \mathtt{b_n}\} \subseteq \Hi\}
\]
The \emph{least fixpoint} of this operator is the least Herbrand model of
$\Pp$ and is the least set of atoms $I$ such that $\Tp(\Hi) \equiv \Hi$.
Let $\Tp^k(\emptyset)$ denote the result of $k$ consecutive calls of $\Tp$, and $\Tp^\infty(\emptyset)$  the least fixpoint interpretation of $\Tp$.
}

\smallskip
\ok{Let $\Ap$ denote the set of all ground atoms that can be constructed from the constants and
predicates occurring in a program~$\Pp$.} A \emph{parameterized interpretation $\Ht$ of  a probabilistic 
program
$\Pp$ }  
 is a set of
tuples $(a,\DNF_a)$ with $a\in\Ap$ and $\DNF_a$ a propositional
formula over  $\Fp$. We say that two parameterized interpretations $\mathcal{I}$
and $\mathcal{J}$ are \emph{equivalent}, $\mathcal{I}\equiv\mathcal{J}$, if
and only if they contain formulas for the same atoms and for all
atoms $a$ with $(a,\varphi)\in\mathcal{I}$ and
$(a,\psi)\in\mathcal{J}$, $\varphi\equiv\psi$.

\ok{Before defining the $\Tcp$ operator for probabilistic programs,
  we introduce some notation. For a parameterized interpretation $\Ht$
  of $\Pp$, we define the set $B(\Ht, \Pp)$  as }
\begin{align*}
B(\Ht, \Pp) = \{(h\theta,\DNF_1\wedge\ldots\wedge \DNF_n) \mid 
\left(h \mathtt{\colonminus  b_1,\ldots,b_n}\right)
    \in \Pp\\ 
\wedge \, h\theta\in\Ap \wedge \forall 1\leq i\leq n: (b_i\theta,\DNF_i)\in\Ht \}
\end{align*}

Intuitively, $B(\Ht, \Pp)$ contains for every grounding of a  rule in
$\Pp$ with head $h\theta$ for which all body atoms have a formula in $\Ht$
the pair consisting of the atom and the conjunction of these formulas.
Note the structural similarity with $\Tp(I)$ above: the $\forall i$ condition \ok{in the definition of $B(\Ht, \Pp)$} corresponds
to the subset condition there, we include substitutions $\theta$ as
our program is non-ground, and we store conjunctions along with the
ground head.

\begin{definition}[$\Tcp$ operator]
\label{def:tcp}
Let 
 $\Ht$ be  a parameterized interpretation of $\Pp$.
Then, the \emph{$\Tcp$ operator is}
$$
\Tcp(\Ht) = \{  (a, \lambda(a,\Ht, \Pp))  \mid a \in \Ap \wedge \lambda(a,\Ht, \Pp)\not\equiv\bot\}
$$
\vspace{-3mm}
where 

\[
\lambda(a,\Ht, \Pp)=
\begin{cases} 
 a &\mbox{if } a\in\Fp \\ 
\bigvee_{(a,\varphi)\in B(\Ht, \Pp)} \varphi & \mbox{if } a\in\Ap\setminus\Fp. 
\end{cases}
\]
\end{definition}

The formula associated with a
derived 
atom $a$ in $\Tcp(\Ht)$ is the disjunction of \ok{$B(\Ht, \Pp)$}  formulae for all
rules with head $a$, but only if this disjunction is not equivalent to
the empty disjunction~$\bot$. The latter is akin to not explicitly listing truth values for
false atoms in regular interpretations.

We have the following correctness results. \ok{$\Tcp^i(\emptyset)$ and
  $\Tcp^\infty(\emptyset)$ are analogously defined  as above.}
\ok{
\begin{theorem}[\citeauthor{vlasselaer:ijar16}~\citeyear{vlasselaer:ijar16}]\label{thm:ijar1}
For a  probabilistic program  $\Pp$, let $\DNF_a^i$ be the formula associated
with atom $a$ in $\Tcp^i(\emptyset)$. For every atom $a\in\Ap$ and total choice $C\subseteq\Fp$,  the following hold:
\begin{compactenum}
\item
For every iteration $i$, we have \[C\models \DNF_a^i \quad \text{ implies } \quad C\cup\Rp\models a\]
\vspace{-2.5mm}

\item
There is an $i_0$ such that for every iteration $i\geq i_0$, we have \[ C\cup \Rp\models a \quad \text{ if and only if } \quad  C\models \DNF_a^i\]
\end{compactenum}
\end{theorem}}

Thus, for every atom $a$, the $\DNF_a^i$ reach a fixpoint
$\DNF_a^{\infty}$ exactly describing the possible worlds entailing
$a$, and the $\Tcp$ operator therefore reaches a
fixpoint where for all atoms $a$, $\Pr(a) = \WMC(\DNF_a^{\infty})$.\footnote{The finite support condition ensures this happens in finite time.}

Algorithm~\ref{alg:tcp} shows how to naively compute $\Tcp(\Ht)$
following the recipe given in the definition. Lines 2--4 compute the set
$B(\Ht,\Pp)$, lines 5--6 add the formulas for  \ok{facts} and 
lines 7--8 the disjunctions that are different from
$\bot$ (i.e., not empty) to the result. \ok{The fixpoint can then  be easily computed, using the equivalence
test as stopping criterion.}

\begin{algorithm}[tb]
\caption{$\Tcp(\Ht)$ for PLP $\Pp$}\label{alg:tcp}
\begin{small}
\begin{algorithmic}[1]
\State $I\defeq \emptyset$; $B\defeq\emptyset$
\For{each $h\colonminus b_1,\ldots,b_n \in\Rp$} 
\For{each $\theta$ with $h\theta\in\Ap \wedge \forall i:
  (b_i\theta,\DNF_i)\in\Ht $}
\State $B \defeq B\cup \{(h\theta, \DNF_1\wedge\ldots\wedge \DNF_n)\}$
\EndFor
\EndFor
\For{each $\ok{f\in\Fp}$}
\State $I\defeq I\cup \{(f,f)\}$
\EndFor
\For{each $a$ with some $(a,\cdot)\in B$}
\State $I\defeq I\cup \{(a, \bigvee_{(a,\varphi)\in B}\varphi)\}$
\EndFor
\State\Return $I$
\end{algorithmic}
\end{small}
\end{algorithm}

The ProbLog2 system provides an implementation of the $\Tcp$ following
\citep{vlasselaer:ijar16}, which proceeds in two steps. It first uses
backward reasoning or SLD resolution to determine the relevant ground
program for the query, i.e., all groundings of rules in $\Rp$ and all
facts in $\Fp$ that
contribute to some derivation of the query, and then iteratively
applies the $\Tcp$ on this program until it reaches the fixpoint  or a
user-provided timeout. In the latter case, the current probability is
reported as lower bound. The implementation of the $\Tcp$  updates
formulas for one atom at a time, using a scheduling heuristic aimed at
quick increases of probability with moderate increase of formula
size.

\medskip \noindent {\bf Magic Sets Transformation.}
 The two most common approaches to logical inference are
\emph{backward reasoning} or \emph{SLD-resolution},
and \emph{forward reasoning}.
The magic sets transformation \citep{bancilhon:pods86,beeri:jlp91} is a well-known technique  to
make forward reasoning query-driven \ok{by simulating backward reasoning}. 
The key idea behind this transformation is to introduce
\emph{magic} predicates for all derived predicates in the program, and
to use these in the bodies of rules as a kind of guard that delays
application of a rule during forward reasoning until the head
predicate of the rule is known to be relevant for answering the
query. To further exploit call patterns, different versions of such
magic predicates can be used for the same original predicate; these
are distinguished by \emph{adornments}. \ok{In this work we exploit  the fact that this transformation preserves query entailment. (More 
details on the transformation are given in the Appendix.)}

\begin{theorem}[\citeauthor{beeri:jlp91}~\citeyear{beeri:jlp91}]\label{thm:magic}
For any set $\Rp$ of rules with non-empty body, any set of facts $\Fp$,
and any query $q$, $\Rp\cup \Fp \models q$ if and only if $\magic{\Rp,q}\cup
\Fp \models q$, where $\magic{\Rp,q}$ is the program obtained by the
magic sets transformation.
\end{theorem}

\section{Semi-naive Evaluation for $\Tcp$}\label{sec:semi}
It is well-known that the computational cost of computing the fixpoint
of the regular $\Tp$ operator can be lowered using semi-naive rather
than naive evaluation~\citep{AbiteboulHV95}. Intuitively, semi-naive evaluation focuses on
efficiently computing the
changes compared to the input interpretation rather than re-computing the full
interpretation from scratch. 
\ok{We now discuss how we apply this idea to the $\Tcp$ operator, where
avoiding redundant work becomes even more important, as we have the added cost of compilation and more expensive
fixpoint checks. }

The high level structure of semi-naive evaluation for $\Tcp$ is given
in Algorithm~\ref{alg:semi}.
\begin{algorithm}[tb]
\caption{Semi-naive fixpoint computation of $\Tcp$ for PLP $\Pp$}\label{alg:semi}
\begin{small}
\begin{algorithmic}[1]
    \State $\Delta\Ht \defeq \{(f,f) \mid f\in\Fp\}$
    \State $\Ht \defeq \Delta\Ht$
    \Repeat
        \State $\Delta\Ht \defeq \Delta\Tcp(\Ht,\Delta\Ht)$
\State $\Ht \defeq \Delta\Ht\cup \{(a,\lambda) \in\Ht\mid 
\neg\exists\lambda': (a,\lambda')\in \Delta\Ht \}$                                                                                                                   	
        \Until{$\Delta\Ht = \emptyset$}                                                                                                                                                
    \State \Return $\Ht$                                                                                                                                             
\end{algorithmic}
\end{small}
\end{algorithm}
We start from the interpretation already containing the
formulas for facts, that is, initially,
the set $\Delta\Ht$ 
of formulas that just changed and the 
interpretation $\Ht$ derived so far contain exactly those pairs. The
main loop then computes the set of pairs for which the formula changes
in line 4, and updates $\Ht$ to contain these new pairs while keeping
the old pairs for atoms whose formulas did not change.  The fixpoint
check in this setting simplifies to checking whether the set of
changed formulas is empty. 

Thus, the task of $\Delta\Tcp(\Ht,\Delta\Ht)$ 
 is to efficiently
compute updated formulas for those derived atoms $a$ whose formula
changes compared to $\Ht$ given $\Delta\Ht$. This is outlined in
Algorithm~\ref{alg:dtcp}. Compared to $\Tcp(\Ht)$ in
Algorithm~\ref{alg:tcp}, line 3 contains an additional condition: we
only add those pairs for which at least one body atom has a changed
formula, i.e., appears in $\Delta\Ht$, that is, the set $D$ is a
subset of the set $B$ in naive evaluation. We no longer need to re-add
fact formulas every time. Where naive evaluation
simply formed all disjunctions from $B$ to add to the result, in
semi-naive evaluation, computing the result is slightly more
involved. For each atom appearing in $D$, we compute the disjunction
over $D$ (line 6). If the atom did not yet have a formula, we add the
disjunction to the output (line 11), otherwise, we disjoin the
disjunction with the old formula and only add this formula to the
output if it is not equivalent to the previous one. Note that this
performs the equivalence check per formula that needs to happen
explicitly in naive evaluation.

\begin{algorithm}[tb]
\caption{$\Delta\Tcp(\Ht,\Delta\Ht)$ for PLP $\Pp$}\label{alg:dtcp}
\begin{small}
\begin{algorithmic}[1]
\State $\Delta\defeq \emptyset$; $D\defeq\emptyset$
\For{each $h\colonminus b_1,\ldots,b_n \in\Rp$} 
\For{each $\theta$ with $h\theta\in\Ap \wedge \forall i:
  (b_i\theta,\DNF_i)\in\Ht $ ~~~~~~~~~~~~~~~$ \wedge 
\exists i:
  (b_i\theta,\DNF_i)\in \Delta\Ht$}
\State $D \defeq D\cup \{(h\theta, \DNF_1\wedge\ldots\wedge \DNF_n)\}$
\EndFor
\EndFor
\For{each $a$ with some $(a,\cdot)\in D$}
\State $\beta_a\defeq \bigvee_{(a,\varphi)\in D}\varphi$
\If{$(a,\lambda_a)\in\Ht$} \Comment{has previous formula $\lambda_a$}
\State $\gamma_a \defeq \lambda_a\vee\beta_a$
\If{$\gamma_a\not\equiv\lambda_a$}  $\Delta\defeq \Delta\cup \{(a,\gamma_a)\}$ \EndIf
\Else \Comment{no previous formula}
\State $\Delta\defeq \Delta\cup \{(a, \beta_a)\}$
\EndIf
\EndFor
\State\Return $\Delta$
\end{algorithmic}
\end{small}
\end{algorithm}

Formally,  
lines 4 and 5 in Algorithm~\ref{alg:semi} implement the following
operator. 
\begin{definition}[$\Scp$ operator]
\label{def:deltatcp}
Let $\Pp$ be a 
probabilistic logic
program.
Let $\Ht$ be a parameterized interpretation of $\Pp$, and
$\Delta\Ht\subseteq\Ht$. 
Let $\mathcal{H}=\Ap\setminus\Fp$. 
Then, the $\Delta\Tcp$ operator is
\begin{align*}
\Delta&\Tcp(\Ht,\Delta\Ht) \\ &= \{(a, \beta_a) \mid a\in\mathcal{H} \wedge 
\neg\exists\lambda : (a,\lambda)\in\Ht \ok{ \, \wedge \beta_a\not\equiv\bot}\}\\
&\cup \{  (a,\lambda\vee\beta_a)  \mid a \in \mathcal{H} \wedge 
   (a,\lambda)\in\Ht \wedge \lambda\not\equiv
  (\lambda\vee \beta_a)\}
\end{align*}
where 
\[
\beta_a=
\bigvee_{(a,\varphi)\in D(\Ht, \Delta\Ht, \Pp)} \varphi \qquad\text{ and }
\]
\begin{align*}
D(\Ht, \Delta\Ht, \Pp) = \{(h\theta,\DNF_1\wedge\ldots\wedge \DNF_n) \mid 
\left(h \mathtt{\colonminus  b_1,\ldots,b_n}\right)
    \in \Pp\\ 
\wedge h\theta\in\Ap \wedge \forall 1\leq i\leq n:
  (b_i\theta,\DNF_i)\in\Ht\\
\wedge\exists 1\leq i\leq n:
  (b_i\theta,\DNF_i)\in \Delta\Ht \}
\end{align*}
The semi-naive $\Tcp$-operator $\Scp$ is 
\begin{align*}
\Scp(\Ht,\Delta\Ht) &=  \Delta\Tcp(\Ht,\Delta\Ht)\\
& \cup \{(a,\lambda) \in\Ht\mid 
\neg\exists\lambda': (a,\lambda')\in \Delta\Tcp(\Ht,\Delta\Ht)\}
\end{align*}
\end{definition}
We are now ready to prove correctness of the approach. 
Let $\Ht^0=\{(f,f)\mid f \in\Fp\}$, 
 let $\Tcp^i(\Ht^0)$ be the result of $i$ consecutive applications
of the $\Tcp$ starting from $\Ht^0$, and 
$\Scp^i(\Ht^0,\Ht^0)$ be the result of $i$ consecutive
applications of the $\Scp$ starting from $\Ht^0=\Delta\Ht^0$ and
using $\Ht^{i+1}=\Scp(\Ht^i,\Delta\Ht^i)$ and $\Delta\Ht^{i+1}=\Delta\Tcp(\Ht^i,\Delta\Ht^i)$. 

The  next claim can be easily proven by induction. 

\smallskip \noindent {\bf Claim } For all $i\geq 1$,
$\Tcp^i(\Ht^0)\equiv\Scp^i(\Ht^0,\Ht^0)$.

\subsection{Semi-naive Fixpoint Computation using Rules}
The key idea behind our efficient implementation of 
the semi-naive fixpoint computation for $\Tcp$ is to introduce
relations that capture the information computed in the steps of 
Algorithms~\ref{alg:dtcp} and~\ref{alg:semi}, and rules that populate
these relations, which can be executed using semi-naive evaluation functionality provided
by an existing Datalog engine, cf.\ Section~\ref{sec:system} below. We discuss the relations and rules
here, abstracting from the specific syntax used in the implementation.

We focus on $\Delta\Tcp$ first,  
\ok{and assume} that, as common in
semi-naive evaluation, the engine ``knows'' which tuples in $\Ht$ are
in $\Delta\Ht$ as well. The input consists of atoms of the form
{\small \verb|lambda(atom,formula)|}, based on which we define three relations
{\small \verb|d(head,conjunction)|} representing the set $D$ of
head-conjunction pairs, {\small \verb|beta(head,disjunction)|}  representing
pairs $(a,\beta_a)$ computed in line 6, and {\small \verb|delta(head,disj)|}
representing the pairs in the output $\Delta$.

The following types of rules populate these predicates. 
For every (potentially non-ground) rule  {\small \verb|h :- b1,...,bn|} in
$\Pp$, we have a rule of the form
\begin{small}
\begin{verbatim}
d(h,conj([F1,...,Fn])) :- 
    lambda(b1,F1), ..., lambda(bn,Fn).
\end{verbatim}
\end{small}
For every predicate {\small \verb|p/n|} defined by rules in $\Pp$, we have 
\begin{small}
\begin{verbatim}
beta(p(X1,...,Xn),disj(L)) :- 
   d(p(X1,...,Xn),_),
   findall(C, d(p(X1,...,Xn),C), L).
\end{verbatim}
\end{small}
Finally, we have the general rules
\begin{small}
\begin{verbatim}
delta(A,D) :- beta(A,D), not lambda(A,_).
delta(A,disj([D,F])) :- 
    beta(A,D), lambda(A,F), 
    not equivalent(disj([D,F]),F).
\end{verbatim}
\end{small}
Here, {\small \verb|findall|} is the usual Prolog predicate that collects all groundings of the first argument for which the second
argument holds in a list and unifies the variable in the third
argument with that list, {\small \verb|not|} is negation as failure, and
{\small \verb|equivalent|} is a special  predicate provided as an external Boolean function that when called on two
ground terms returns true if the propositional formulas encoded by
these terms are logically equivalent.

When computing the fixpoint, the relation {\small \verb|delta|}
provides the next $\Delta\Ht$, and we need to compute the
next $\Ht$ from the current {\small \verb|lambda|} and the {\small \verb|delta|} as in
line 5 of Algorithm~\ref{alg:semi}. We next show how we integrate this
step into the rules. The
key idea is to extend the {\small \verb|lambda|} predicate with an additional
argument that contains a unique identifier for every atom-formula pair
added to the relation,  to mark pairs that contain outdated
formulas for an atom by adding the identifier to a new relation
{\small \verb|outdated|}, and to incrementally populate these two relations
across the fixpoint computation while dropping all intermediate
relations after each iteration. 

The most recent formula for an atom \verb|a| is now given by the
conjunctive query 
{\small \verb|lambda(a,F,I), not outdated(I)|.}  We adapt the rules defining
{\small \verb|d|} to use this pattern in the body, keep the rules for
{\small \verb|beta|} unchanged, and replace the rules for
{\small \verb|delta|} with the following set of rules:
\begin{small}
\begin{verbatim}
lambda(A,D,A) :- 
  beta(A,D), not lambda(A,_,_).
aux(A,disj([D,F]),I) :-
  beta(A,D), lambda(A,F,I), not outdated(I),
  not equivalent(disj([D,F]),F).
lambda(A,F,u(I)) :- aux(A,F,I).
outdated(I) :- aux(A,F,I).
\end{verbatim}
\end{small}
The first of these covers the case where atom {\small \verb|A|} did not have a
formula yet, in which case we use the atom itself as identifier. The
second rule defines an auxiliary relation whose elements are an atom,
the updated formula for the atom, and the identifier for the atom's
previous formula, where the latter is used in the last two rules to
generate a new identifier and mark the old one as outdated. 

After reaching the fixpoint, we perform a final projection step to eliminate identifiers using the following rule
{\small \verb|lambda(A,F) :- lambda(A,F,I), not outdated(I).|}

\section{Magic Sets for $\Tcp$}\label{sec:magic}
So far, we have discussed how to efficiently compute the full fixpoint
of the $\Tcp$ operator. However, in practice, we are often only
interested in specific queries. In this case, regular $\Tp$ often uses
the magic set transformation to restrict the fixpoint computation to
the atoms relevant to the query. We now show that we can apply the same
transformation in our setting to make the $\Tcp$ goal-directed, and
then introduce an optimization of our approach for magic programs.

\ok{We fix a probabilistic program $\Pp=(\Rp, \Fp, \pi)$}.
\begin{definition}[Magic Sets for PLP]\label{def:mplp}
The \emph{magic transform of  $\Pp$ with
respect to a query $q$} is the program $\Mp = (\mathsf{magic}(\Rp,q), \Fp, \pi)$, where
$\mathsf{magic}(\cdot,\cdot)$ \ok{is as in Theorem~\ref{thm:magic}.}
\end{definition}
That is, we apply the regular magic set transformation to the rules
with non-empty body and keep the facts and labeling.

\begin{theorem}\label{thm:plainmagic}
\ok{Let 
$q$ be a query}. 
The formula $\lambda(\Mp,q)$ associated with $q$ in the
fixpoint of $\Tcm$ 
is equivalent to the formula $\lambda(\Pp,q)$ associated with $q$ in the
fixpoint of $\Tcp$. 
\end{theorem}
This is a direct consequence of Theorems~\ref{thm:ijar1} (Point~2)
and~\ref{thm:magic}.

While the above 
makes our approach
query directed, compiling formulas for magic atoms may introduce
significant overhead. We therefore now define an optimized version of
\ok{goal-directed} $\Tcp$ that avoids compilation for magic atoms, and show that this
computes correct formulas.

\begin{definition}[magic-$\Tcp$ operator]
\label{def:mcp}
Let 
$q\in\Ap$ be the query of interest, and $\Ht$ 
a parameterized interpretation of the magic transform $\Mp$ \ok{of $\Pp$}. 
Then, the magic-$\Tcp$ operator \ok{$\Mcp(\Ht)$} is defined as
$$
 \{  (a, \mu(a,\Ht, \Mp))  \mid a \in \MAp \wedge \mu(a,\Ht, \Mp)\not\equiv\bot\}
$$
where 
\[
\mu(a,\Ht, \Mp)=
\begin{cases} 
\lambda(a,\Ht, \Mp) & \mbox{if }a\in\Ap \\
 \top &\mbox{if "magic case"} 
\end{cases}
\]
where $\lambda$ is as in Definition~\ref{def:tcp} above, and "magic case"
means 
that $a\in\MAp\setminus\Ap$ and there is a rule $h\colonminus b_1,\ldots,b_n$ in
$\Mp$ and a grounding substitution $\theta$ such
that $a=h\theta$ and for each $b_i$ there is a $\lambda_i$ such that
$(b_i\theta,\lambda_i)\in\Ht$. 
\end{definition}
That is, if $a$ is not a magic atom (which includes the \ok{facts}),
we use the same update operations as for the regular \ok{$\Tcm$} 
operator with the magic program, but for magic atoms, we set the formula $\top$ if (and only if) there is a rule that can derive the atom from the current interpretation $\Ht$.

\begin{theorem}\label{thm:topmagic}
\ok{For
a query $q$}, let $\tau_q^i$ be the formula associated
with $q$ in $\Mcp^i(\emptyset)$. For every total
choice $C\subseteq\Fp$,
there is an $i_0$ such that for every iteration $i\geq i_0$,  $\tau_q^i$ exists and 
\vspace{-2mm}
 \[ C\cup \Rp\models q \quad \ok{\text{if and only if}} \quad  C\models \tau_q^i\]
\end{theorem}
\ok{The proof relies on two intermediate results: Formulas computed by $\Tcp$ are always lower bounds for those computed by $\Mcp$; and for any atom in the original program (i.e., excluding the magic atoms), the formulas computed by $\Mcp$ only include correct choices relative to the original program.  These combined with Theorem \ref{thm:ijar1} (Point 2) provide us with the desired result, cf.\ Appendix.
}

\section{Implementation}\label{sec:system}

We use VLog \citep{urbani:aaai16,CDGJKU2019} to implement the principles introduced in Section~\ref{sec:semi}, and refer to this implementation, which we will make
available upon publication of the paper, as \vp.
We use VLog
because it has been shown to be efficient, 
is open source,
supports negation (which is used by some of our transformed rules) and
provides very efficient rule execution functions that we can invoke directly.

As VLog is a Datalog engine, we cannot directly use the rules as
introduced in Section~\ref{sec:semi}. Instead, we encode functors into
predicate or constant names whereever possible and use a procedural
variant of the second set of rules to avoid findall. 
We implement the ``equivalent'' function as an external function
  over Sentential decision diagrams (SDD) \citep{darwiche:ijcai11},
  using the SDD package
developed at
UCLA\footnote{\label{fn:ucla}\texttt{http://reasoning.cs.ucla.edu/sdd/}}.

More precisely, given program $\Pp$, query $q$ and iteration parameter $d$, \vp\ performes the following steps:
\begin{compactenum}
\item Apply the magic set transformation to $\Rp$ and $q$. 
\item Partition $\mathsf{magic}(\Rp,q)$ into three sets, where $\Rp_1$ contains the rules
  whose bodies only use facts, $\Rp_2$ contains all other rules that do not depend on cyclic derivations, and $\Rp_3$ the remaining ones, and
  apply the transformation of Section~\ref{sec:semi} to all three.
\item Materialize the fixpoint, i.e., execute the transforms of $\Rp_1$
  and $\Rp_2$ in sequence, iteratively execute the transform of $\Rp_3$
  $d$ times (or until fixpoint if $d=\infty$), and finally use the projection rule to eliminate
  bookkeeping identifiers.
\item For each query answer in the materialization, compute the WMC of
  its SDD.
\end{compactenum}
This gives us \vpplain, an implementation of~$\Tcm$, i.e., without the
optimizations introduced in  Section~\ref{sec:magic}.
To implement the modified operator~$\Mcp$, we note
that instead of explicitly setting the formula of a magic atom $\top$ in
our transformed program, we can simply not store magic atoms in the \verb|lambda|
relation but instead keep them as usual whereever they appear. We
refer to this as \vpopt.

\section{Experimental Evaluation}\label{sec:exp}
We perform experiments using the two versions of \vp\ as well as the
implementation  of $\Tcp$ provided by the 
ProbLog2
system\footnote{\texttt{https://dtai.cs.kuleuven.be/problog/}, version
2.1.0.37, option \texttt{-k fsdd}}, which we refer to as \problog, to
answer the following questions: 
\begin{compactdesc}
\item[Q1] Does \vpopt\ outperform \vpplain\ in terms of
  running times and scalability?
\item[Q2] How does the performance of \vpopt\ compare to that  of
  \problog?
\item[Q3] How scalable is \vpopt?
\end{compactdesc}

As \vp\ aims to improve the reasoning phase, but essentially keeps the
knowledge compilation phase unchanged, we distinguish two types of
benchmarks: those where the bottleneck is reasoning, i.e., propositional formulas are relatively
small, but a large amount of reasoning may be required to identify the formulas, and those where
the bottleneck is knowledge compilation, i.e., propositional formulas
become complex quickly.  
As a representative of the former, we use a probabilistic version of
\lubm~\citep{lubm}, a setting known to be challenging for
ProbLog~\citep{schoenfisch:ijar17,vBDJ-CIKM19}; for the latter type, we  use
three benchmarks from the ProbLog
literature~\citep{fierens:tplp15,renkens:aaai14,vlasselaer:ijar16} that
essentially are all variations of network connectivity queries:
\begin{description}
\item[LUBM] We create a probabilistic version of the \lubm\ benchmark
 by adding a random probability from $[0.01,1.0]$ to each
  fact in the database, using three databases of increasing size
  (approximately 19K
  tuples in \lubmS, 1M tuples in \lubmM\ and 12M tuples in \lubmL). We drop all rules that introduce existentials,
  as these are not supported in our setting, and use the 14 standard
  queries (with default join order, cf.~Appendix).
\item[WebKB] We use the
\webkb\footnote{\texttt{http://www.cs.cmu.edu/webkb/}} dataset restricted to the 100 most frequent words \citep{davis:icml09}
and with random probabilities from \big[0.01, 0.1\big], using all
pages from the Cornell database. This results in a dataset with 63
ground 
queries. 
\item[Smokers] We
  use random  power law graphs with increasing numbers of persons for
  the standard `Smokers' social network domain, and non-ground query
  \verb|asthma(_)|. 
\item[Genes] We use the biological network
of \citeauthor{ourfali:bioinf07}~\citeyear{ourfali:bioinf07} and 50 of its connection queries on gene
pairs. 
\end{description}

To answer \textbf{Q1}, we compare running times of both versions of
\vp\ on \lubmS. The first two blocks of Table~\ref{tab:lubmlong}
report times for materialization and WMC as well as their sum,
with a two minute timeout on the former. The differences in
materialization time clearly show that compiling formulas for magic
atoms can introduce significant overhead and should thus be
avoided. 
We thus answer \textbf{Q1} affirmatively, and only use
\vpopt\ (or \vp\ for short) below.

\begin{table*}
\caption{Results for \lubm: time for materialization, weighted model
  counting and total time for \vpplain\ (two minute timeout for materialisation) and \vpopt\ on \lubmS\ (Q1),
  grounding time, compilation time and total time for \problog\ (two minute overall
  timeout, marked x)   as well as type of answer (exact or lower bound) provided if any (Q2), and times for \vpopt\ on \lubmM\ and \lubmL\ (Q3). All
times in seconds. }
\label{tab:lubmlong}
\footnotesize
\begin{tabular}{l|rrr|rrr|rrrl|rrr|rrr}
& \multicolumn{10}{c|}{\lubmS} & \multicolumn{3}{c|}{\lubmM} & \multicolumn{3}{c}{\lubmL}\\
&\multicolumn{3}{|c|}{\vpplain} & \multicolumn{3}{c|}{\vpopt} & \multicolumn{4}{c|}{\problog} & \multicolumn{3}{c|}{\vpopt} & \multicolumn{3}{c}{\vpopt} \\
	&	mat 	&	wmc 	&	total&	mat	&	wmc	&	total	&	ground	& compile &	total 	&	answer&	mat	&	wmc	&	total&	mat	&	wmc	&	total	\\\hline
q01	&	0.1	&	0.1	&	0.1	&	0.0	&	0.1	&	0.1	&	0.7	& 1.2 &	40.9	&	exact &	0.5 &	0.2 &	0.7 &	6.7 &	1.8 &	8.5	\\
q02	&	0.5	&	0.0	&	0.5	&	0.1	&	0.0	&	0.1	&	x	&&		&	-&	6.9 &	0.6 &	7.5 &	&&			\\
q03	&	0.1	&	0.1	&	0.2	&	0.0	&	0.1	&	0.2	&	6.3	&	x&	&	exact &	0.2 &	0.3 &	0.5 &	2.3 &	1.9 &	4.1	\\
q04	&	x	&		&	x	&	3.6	&	0.5	&	4.2	&	49.6	&	x	&&	-&	5.2 &	0.7 &	5.8 &	&&		\\
q05	&	x	&		&	x	&	6.3	&	11.3	&	17.5	&	56.4	&	x&	&	-&	7.5 &	9.8 &	17.3 &	&&		\\
q06	&	x	&		&	x	&	41.8	&	113.4	&	155.2	&	47.7	&	x&	&	- &					&&&&&	\\
q07	&	6.3	&	0.9	&	7.3	&	3.4	&	1.0	&	4.4	&	x	&&		&	-&	5.1 &	1.1 &	6.2 &		&&	\\
q08	&	x	&		&	x	&	60.6	&	115.2	&	175.8	&	x&	&		&	- &					&&&&&	\\
q09	&	x	&		&	x	&	22.2	&	2.8	&	25.0	&	x	&&		&	- &					&&&&&	\\
q10	&	1.6	&	0.1	&	1.6	&	0.5	&	0.1	&	0.6	&	50.6	&	x	&&	bound &	1.2 &	0.2	& 1.5	&&&	\\
q11	&	0.5	&	5.3	&	5.8	&	0.3	&	3.4	&	3.7	&	0.4	&0.9&	40.7	&	exact &	0.3 &	3.5 &	3.8 &	0.4 &	4.9 &	5.3	\\
q12	&	x	&		&	x	&	15.5	&	0.2	&	15.7	&	47.2	&	x&	&	-&	17.9 &	0.4 &	18.3&&&		\\
q13	&	2.1	&	0.0	&	2.1	&	1.0	&	0.0	&	1.1	&	57.9	&	x	&&	bound	& 12.7 &	0.6 &	13.3	&&&	\\
q14	&	0.2	&	80.2	&	80.4	&	0.2	&	89.4	&	89.6	&	1.2	& 0.2 &	87.7	&	exact						&3.1&&&&&
\end{tabular}
\end{table*}
To answer \textbf{Q2}, we consider all benchmarks, with a two minute timeout 
per query for \problog. 
In the middle of Table~\ref{tab:lubmlong}, we list the times \problog\
reports for grounding, compiling SDDs, and total time
(which also includes data loading) per query on \lubmS. 
The \emph{answer} column indicates whether \problog\
returns exact probabilities (fixpoint detected),
lower bounds (timeout with partial formula for query, before detecting
fixpoint) or no result (timeout without formula for query). The lower bounds reported for q10 and q13 are practically the final probabilities,
but \problog\ has not detected this yet at timeout. \problog\
reaches the timeout during grounding for q02, q07, q08 and q09, for
which the default join order first builds a Cartesian product of two
or three type relations, which is the worst possible join order for
SLD resolution. 

\ok{Comparing \problog's times to the times for \vpopt\ to the left, we observe
that \vp\ materializes fixpoints, which includes compiling formulas
(using the same SDD tool) in often significantly less time than
it takes \problog\ to just determine the relevant ground program with
the exception being the database lookup query q14, where running times
are similar. }
These results clearly demonstrate the benefits of
exploiting Datalog techniques in terms of speed.

On the PLP benchmarks, \problog\ can only compute lower bounds for
most queries, and \vp\ cannot fully materialize the fixpoint,
as SDDs quickly become too large to handle. For
\vp, we
therefore select for each benchmark a fixed number of iterations close
to the feasibility borderline. Intuitively, this restricts the length of paths
explored, though that length does not equal the number of iterations
due to our program transformations, and \vp\ thus always computes lower
bounds. Here, we are thus interested to see whether \vp\ can achieve
comparable or better lower bounds to \problog\ in less time. 

For \webkb, seven of the 63 queries are easy, as the corresponding
pages have incoming link chains of length at most two, whereas
formulas explode for all others. 
\problog\ computes probabilities for the easy queries in at most seven
seconds, and bounds after timeout for the others, whereas \vp\
computes bounds using four iterations for all queries in at most seven
seconds. 
All differences between lower bounds are smaller than $0.02$, 
with \problog\ achieving higher bounds on 47
queries and smaller bounds on 11.

For \smokers, \problog\ quickly computes exact answers for all networks up to
size 12, and reaches the timeout on all networks from size 19
onwards. We therefore consider 10 networks for each size 
from 10 to 20 persons, using three iterations for \vp. 
Table~\ref{tab:smokers} lists for each size the number of scenarios
(out of ten) solved exactly by \problog, along with minimum, average
and maximum total times for both systems (excluding timeout cases). We
again observe a clear time advantage for \vp\ with increasing
size. Furthermore, the bounds provided by \vp\ are close to actual
probabilities (exact for 89 queries, at most $0.002$ lower for 615
queries) where \problog\ computes those, and close to \problog's lower
bounds otherwise (up to $0.05$ higher on 570 queries, and at most
$0.002$ lower for 312). 

\begin{table}
\caption{\smokers: \problog: number of scenarios (of 10) solved exactly
  in 2 minutes, min/avg/max times in seconds over those;
  \vp: min/avg/max times in seconds over all neworks to compute lower
  bounds with 3 iterations}
\label{tab:smokers}
\small
\begin{tabular}{r||r|r|r|r||r|r|r}
& \multicolumn{4}{c||}{\problog} & \multicolumn{3}{c}{\vp}\\
size & exact& min & avg & max & min & avg & max \\\hline
10 &	10 &	0.1	& 0.6 &	3.3  &	0.5	&	0.6	&	1.0\\
11 &	10 &	0.1	& 1.0 &	5.7  &	0.5	&	0.6	&	0.8\\
12 &	10 &	0.2	& 10.9 &	93.1 &	0.6	&	0.8	&	0.9 \\
13 &	9 &	0.2 &	15.7 &	55.9  &	0.6	&	0.8	&	1.2\\
14 &	8 &	0.7	& 34.9 &	106.7 &	0.3	&	1.2	&	3.6 \\
15 &	3 &	2.7	& 34.2 &	65.6  &	0.8	&	1.3	&	2.5\\
16 &	3 &	12.0 &	24.6 &	41.8&	0.7	&	1.6	&	3.3  \\
17 &	2 &	1.7	& 8.4	& 15.1  &	0.8	&	2.1	&	4.5\\
18 &	1 &	23.6 &	23.6	& 23.6 &	0.9	&	2.7	&	5.8 \\
19 &	0 &		&&	  &	1.1	&	4.6	&	12.1\\
20 &	0 &		&&	 &	1.3	&	3.9	&	13.4
\end{tabular}
\end{table}

For \genes, \problog\ reaches the timeout and thus computes lower
bounds for all queries. As those range from $0.0108$ to $0.9999$, we
run \vp\ for 1, 3, 5, 6 and 7 iterations to explore the effect of the
approximation in more detail. 5 iterations are infeasible for two
queries, 7 iterations for another five queries. \vp\ running times for the last feasible iteration of each
query are below five seconds per query for all but two queries, which
take 13 and 78 seconds, respectively. 
We refer to the first non-zero bound for
query $q$ 
\vp\ reaches  as $f_q$, to the last (feasible) one as
$l_q$, and to \problog's bound as $p_q$. We observe $p_q<f_q$ for 23
queries, $f_q\leq p_q<l_q$ for 15, and $l_q\leq p_q$ for 12. The
actual differences $p_q-l_q$ vary from $-0.88$ to $0.20$, indicating
that this is a diverse set of queries where no single
approximation strategy suits all, but also that \vp\ often achieves
much higher bounds in less time.

Based on these results, as an answer to \textbf{Q2}, we conclude that
\vp\ often speeds up logical inference time significantly, which
enables inference in scenarios that have previously been infeasible
due to the grounding bottleneck. At the same time, for benchmarks
where the bottleneck is knowledge compilation, its use of magic
sets provides a natural
scheduling strategy for formula updates that achieves similar bounds
to those of \problog, but often in less time.

To answer \textbf{Q3}, we consider the three \lubm\
databases. Table~\ref{tab:lubmlong} lists running times of \vpopt\ (for
materialization, weighted model counting, and total of both) without
iteration limit per query in
seconds. Blank entries indicate that the corresponding phase did not
finish due to problems with SDDs. 
The three queries that are answered on the largest database all have
one element answers from relatively narrow classes with a direct link
to a specific constant, which limits both the number of answers and
the size of formulas. In contrast, the three queries that cannot be
solved on the medium size database either ask for a broad class (all
students, q06) or triples that include members of such a class and two
related objects (q08 and q09), which means both more 
answers on larger databases and more complex SDDs.

Taking all results together, our answer to \textbf{Q3} is that
\vp\ directly benefits from the scalability of VLog  for logical
reasoning, but \ok{as all WMC-based approaches  is limited by the complexity of
formula manipulation}.

\section{Related Work}\label{sec:related}
A related formalism is  the  probabilistic version of 
Datalog 
by  \citeauthor{BaranyCKOV17}~\citeyear{BaranyCKOV17}, but it does not  adopt the distribution semantics and there is no existing system. pOBDA can be also viewed as a variation of PLP formalisms~\citep{JungL12}. 
The exact relation to PLP broadly depends on the   ontology language of choice, usually based on description logics or existential rules, see~\citep{BorgwardtCL18} for a recent survey. 

Finally, we note that magic sets  have been  also applied to other extensions of Datalog, e.g.\ with aggregates, equality or disjunctive Datalog~\citep{DBLP:conf/lpnmr/AlvianoGL11,BenediktMT18,DBLP:journals/ai/AlvianoFGL12}.

\section{Conclusions}\label{sec:conc}
We have adapted and extended the well-known Datalog techniques of semi-naive
evaluation and magic sets to avoid the grounding bottleneck of
state-of-the-art inference in probabilistic logic programming,
contributed a prototype implementation based on VLog,
and experimentally demonstrated the benefits in terms of scalability
on both traditional PLP benchmarks and a query answering scenario that
previously has been out of reach for ProbLog.
Immediate future work includes extending the  approach to stratified
negation 
and eliminating the need
to pre-determine the depth parameter by iterative expansion, which
will provide a proper anytime algorithm. Beyond this,
we intend to perform experiments on additional benchmarks in the knowledge graph and
ontology setting, \ok{including ontologies based on description logics and existential rules},
to study ways to support probabilistic programs with
functors, 
as well as alternative ways to trade-off time spent on logical reasoning
vs compilation, e.g., by restricting the number of fixpoint checks performed.

\bibliographystyle{plainnat}
\bibliography{references}
\clearpage
\section*{Appendix}\label{sec:appendix}

\subsection{Additional Background}

\noindent {\bf Logic Programming.}
In this work, we only consider function-free logic programs. 
A \emph{term} is a variable or a constant.
An \emph{atom} is of the form $p(t_1,\dots,t_m)$ where $p$ is a \emph{predicate} of \emph{arity} $m$ and   $t_i$ is a term.
A  \emph{rule} (or \emph{definite clause}) is a universally quantified expression of the form $\mathtt{h \colonminus b_1, ... , b_n}$ where
$\mathtt{h}$ and the $\mathtt{b_i}$ are atoms and the comma denotes conjunction.
We call  $\mathtt{h}$ the  \emph{head} of the rule and
$\mathtt{b_1, ... , b_n}$ the \emph{body}. Intuitively, a rule states that 
whenever the body is true, the head has to be
true as well. 
We assume the standard wellformedness condition:  all logical variables in the  head of a rule
also appear in its body.
A \emph{fact} is a rule with $n=0$ and is written more compactly as $\mathtt{h}$.
A \emph{logic program} (or \emph{program} for short) is a finite set of rules.
For ease of presentation, throughout the paper we adopted the equivalent database-view
of logic programs, i.e., a program $P$ has two components: a set $\Rp$ of non-fact rules
 and a set of  facts $\Fp$, called \emph{database}. Clearly, $\Rp \cup \Fp$ is a logic program.

\smallskip
An expression is \emph{ground} if  it does not contain variables. Let $\Ap$ be the set of all ground atoms that can be constructed from the constants and
predicates occurring in a program~$\Pp$.  A \emph{Herbrand interpretation}
of $\Pp$ is a truth value assignment to all  $a\in\Ap$, and is often
written as the subset of $\mathtt{true}$ atoms (with all others being
$\mathtt{false}$), or as a conjunction of atoms. A Herbrand
interpretation satisfying all rules in the program $\Pp$ is a \emph{Herbrand
model of $\Pp$}. The model-theoretic semantics of a program is given by its unique \emph{least Herbrand model},
that is, the set of all ground atoms $a\in\Ap$ that are entailed by
the logic program, written $\Pp\models a$. 

The task of logical inference is to determine whether a program $\Pp$
entails a given atom, called  \emph{query}. 

\medskip \noindent {\bf $\Tcp$. }
Algorithm~\ref{alg:naive} computes the fixpoint, where the
equivalence test used as stopping criterion is as defined above. It
starts from the interpretation already containing the formulas for
\ok{facts} to simplify comparison with the semi-naive version
introduced in Section~\ref{sec:semi}.

\begin{algorithm}[h!]
\caption{Naive fixpoint computation of $\Tcp$ for PLP $\Pp$}\label{alg:naive}
\begin{small}
\begin{algorithmic}[1]
     \State $\Ht \defeq \{(f,f) \mid \ok{f\in\Fp}\}$
    \Repeat
\State $J\defeq\Ht$
        \State $\Ht \defeq \Tcp(J)$
        \Until{$\Ht \equiv J$}                                                                                                                                                
    \State \Return $\Ht$                                                                                                                                             
\end{algorithmic}
\end{small}
\end{algorithm}

\medskip  \noindent {\bf Magic Sets. } We focus on aspects necessary to
our work and refer to \citep{bancilhon:pods86,beeri:jlp91} for full details and the general
case.

\smallskip \noindent 
{\it Adornments.} 
    An \emph{adornment} for an $n$-ary predicate $p$ is a
    string $\alpha$ of length $n$ over alphabet $\ad{b}$ (``bound'') and
    $\ad{f}$ (``free''), and $\mgc{p}{\alpha}$ is a fresh \emph{magic}
    predicate unique for $p$ and $\alpha$ with arity equal to the number of
    $\ad{b}$-symbols in $\alpha$. 
    For $\alpha$ an adornment of length $n$
    and ${\vec t}$ an $n$-tuple of terms, ${\vec t^\alpha}$ contains in the
    same relative order each ${t_i \in \vec t}$ for which the $i$-th element of
    $\alpha$ is $\ad{b}$.
 Let $\mathsf{vars}(t)$ denote the variables occurring  in $t$. For ${L(t_1, \dots, t_k)}$ a literal and $V$ a set of variables,
    ${\mathsf{adorn}(L(t_1, \dots, t_k),V)}$ returns an adornment $\alpha$ for
    \ok{the predicate of $L$} such that  the variables of $\mathsf{vars}(t_j)\subseteq V$  if the $j$-th element of $\alpha$
    is $\ad{b}$. In our implementation, we bound as much as possible
    and also use adornment subsumption. 
While general versions of magic sets also employ a literal reordering
function, we omit this for simplicity as we use identity in our implementation. 

\smallskip
Algorithm~\ref{alg:magic} below defines the magic set transformation we use
here. Without loss of generality, we assume that there is a single 
query predicate $q$  of arity $n$ which is queried with distinct logical
variables for each argument. Line 1 initializes the queue $\mathcal{T}$ of magic
predicates to be defined and the set $\mathcal{D}$ of magic predicates
considered with $\mgc{q}{\ad{f} \cdots \ad{f}}$,
where the length of the adornment is the arity of the query
predicate. The main loop then processes the magic predicates in
$\mathcal{T}$ in turn, looping over the rules defining the
corresponding original predicate. Line 5 adds a version of the rule
extended with the magic guard to the program $\mathcal{S}$. Lines 6 to
9 generate the relevant magic predicates for those body atoms in the
rule whose predicates are defined by rules in $P$. Line 7 generates an
adornment based on the variables in the head and the preceding body
atoms, line 8 adds a rule to the program stating that the $i$th atom's
predicate is relevant if the head's predicate is relevant and the
first $i-1$ body atoms are true, and line 9 schedules the magic
predicate to be processed if it has not been scheduled yet. 

\begin{algorithm}[h!]
\caption{$\magic{P,q}$}\label{alg:magic}
\begin{small}
\begin{algorithmic}[1]
    \State $\mathcal{D} \defeq \mathcal{T} \defeq \{ \mgc{\predQ}{\ad{f} \cdots \ad{f}} \}$ and $\mathcal{S} \defeq \emptyset$ 				\label{alg:magic:init}
    \While{$\mathcal{T} \neq \emptyset$}
        \State \textbf{choose and remove} some $\mgc{p}{\alpha}$ from $\mathcal{T}$                                                                                                 											\label{alg:magic:R:start}
        \For{\textbf{each} rule $p(\vec t)\colonminus b_1,\ldots,b_n\in P$}                                                                                             																\label{alg:magic:r:start}
        \State \textbf{add} $p(\vec t)\colonminus \mgc{p}{\alpha}(\vec
        t^\alpha) , b_1,\ldots,b_n $ to $\mathcal{S}$                                                                    									\label{alg:magic:mod-rule}
         \For{each $b_i = p_i(\cdot)$ with a $p_i(\vec x)
          \colonminus \ldots \in P$}                                                                                              													\label{alg:magic:body:start}
            \State $\gamma \defeq \mathsf{adorn}(b_i,\vars{\vec t^\alpha} \cup \bigcup_{j=1}^{i-1} \vars{b_j})$                                                       												\label{alg:magic:adorn}
            \State \textbf{add} $\mgc{p_i}{\gamma}(\vec t_i^\gamma)
            \colonminus \mgc{p}{\alpha}(\vec t^\alpha) , b_1,\ldots,b_{i-1}$ to $\mathcal{S}$                           			\label{alg:magic:magic-rule}
            \If{$\mgc{p_i}{\gamma} \not \in \mathcal{D}$}
                \textbf{add} $\mgc{p_i}{\gamma}$ to $\mathcal{T}$ and $\mathcal{D}$                                                                                                                 									\label{alg:magic:add-S}
            \EndIf
        \EndFor                                                                                                                                                                     																		\label{alg:magic:body:end}
        \EndFor                                                                                                                                                                     																		\label{alg:magic:r:end}
    \EndWhile                                                                                                                                                                       																	\label{alg:magic:R:end}
    \State \Return $\mathcal{S}$                                                                                                                                                   	 															\label{alg:magic:return}
 \end{algorithmic}
\end{small}
\end{algorithm}

\subsection{Proofs for Section~\ref{sec:semi}}
\paragraph{Claim: } For all $i\geq 1$, $\Tcp^i(\Ht^0)\equiv\Scp^i(\Ht^0,\Ht^0)$.

\smallskip \noindent {\bf Proof sketch: }
For a fact $f \in \Fp$, the $i$th iteration
of $\Tcp$ re-adds
$(f,f)$ and $\Scp$ copies $(f,f)$ from $\Ht^{i-1}$, as $\Delta\Tcp$ never
updates these. 
For derived atoms that do not
have a formula in $\Ht^{i-1}$,  $B(\Ht^{i-1},\Pp)$ and $D(\Ht^{i-1},\Delta\Ht^{i-1},\Pp)$
contain the same conjunctions, and both operators therefore add the same
formula to their respective result. 
Let $a$ be a derived atom with
$(a,\lambda)\in\Ht^{i-1}$. 
Note that  $D^i\subseteq B^i$ and $B^i\setminus D^i \subseteq
B^{i-1}$, i.e., every element of $B^i$ either appears in $D^i$ (if
some body atom has a formula in $\Delta\Ht^{i-1}$ or already appeared
in the previous round's $B^{i-1}$. 
If all conjunctions for $a$ appear in $D^i$, both algorithms construct
the same formula. If all conjunctions for $a$ appear in $B^i\setminus
D^i$, $\Tcp$ reconstructs the same formula for $a$ as in the previous
iteration, and $\Scp$ copies the formula from the previous iteration,
so again both are identical. Let $a$ be an atom with conjunctions in
both subsets. In this case, $\Tcp$ directly computes the disjunction
over both subsets, whereas $\Scp$ computes the disjunction over $D$
only, disjoins this with the $\lambda$ from the previous round (which
includes the formula over $B^i\setminus D^i$ as a disjunct), and
then decides whether to update or keep the previous formula,
i.e., we have the formula for $a$ in $\Ht^{i-1}$, which is $\lambda^{i-1} =
\phi\vee\psi$, the new formula computed by $\Tcp$, which is
$\lambda^i=\beta^i\vee\psi$ (i.e., $\psi$ is the disjunction over the
rules whose bodies did not change), and the new formula computed by $\Scp$,
which is $\lambda^{i-1}\vee\beta^i = \phi\vee\psi\vee\beta^i$ (either
returned as new formula in $\Delta\Tcp$ if it is different from $\lambda^{i-1}$, or in its equivalent form
$\lambda^{i-1}$ by copying). So we need to show that
$\beta^i\vee\psi\equiv \phi\vee\psi\vee\beta^i$. As our program is
monotone, once a conjunction for a rule body appears in some $B$, a
conjunction for that rule body will appear in all subsequent $B$s as
well. Thus, for any conjunction $c$ in $\phi$, there is an updated
conjunction $c'$ in $\beta^i$ whose models include those of $c$, and
$\phi$ is thus redundant in $\phi\vee\psi\vee\beta^i$. QED

\subsection*{Proofs for Section~\ref{sec:magic}}
{\bf Theorem~\ref{thm:plainmagic}} {\it  Let 
$q$ be a query. 
The formula $\lambda(\Mp,q)$ associated with $q$ in the
fixpoint of $\Tcm$ 
is equivalent to the formula $\lambda(\Pp,q)$ associated with $q$ in the
fixpoint of $\Tcp$. 
}

\smallskip \noindent {\bf Proof sketch}
As both $\Pp$ and $\Mp$ are PLPs, by Theorem~\ref{thm:ijar1} Point~2, we have
\begin{align*}
\lambda(\Mp,q)&\equiv \bigvee\limits_{C\subseteq \Fp:   C\cup  \RMp\models q}\, \bigwedge_{ f\in   C}f\\
\lambda(\Pp,q)&\equiv \bigvee\limits_{C\subseteq \Fp:   C\cup       \Rp \models q}\, \bigwedge_{\ok{f}\in     C} f
\end{align*}
and every total choice $C\subseteq\Fp$ is a  
database. As the programs are functor-free, by Theorem \ref{thm:magic}, for any such $C$,
$C\cup \mathsf{magic}(\Rp,q)  \models q$ if and only if $C\cup\Rp  \models q$, i.e., the selection criteria are equivalent. Thus, $\lambda(M,q)\equiv\lambda(P,q)$. QED.

\medskip
\noindent {\bf Theorem~\ref{thm:topmagic}} {\it For 
a query $q$, let $\tau_q^i$ be the formula associated
with $q$ in $\Mcp^i(\emptyset)$. For every total
choice $C\subseteq\Fp$,
there is an $i_0$ such that for every iteration $i\geq i_0$,  $\tau_q^i$ exists and 
 \[ C\cup \Rp\models q \quad \ok{\text{if and only if}} \quad  C\models \tau_q^i\]}

\smallskip \noindent {\bf Proof sketch. }
 The proof relies on two following intermediate results, which are proven below.
We first show that the formulas computed by $\Tcp$ are always lower bounds for those computed by $\Mcp$.
\begin{lemma}\label{lemma:claim4}
\ok{For 
an atom $a\in\MAp$}, let $\tau_a^i$ be the formula associated
with $a$ in $\Mcp^i(\emptyset)$ and $\delta_a^i$ be the formula associated
with $a$ in $\Tcm^i(\emptyset)$. For each iteration $i$ and each atom $a$, $\tau_a^i$ exists if and only if $\delta_a^i$ exists. If they exist, for every total choice $C\subseteq\Fp$, 
$C\models \delta_a^i \ok{\text{ implies that }} C\models\tau_a^i$.
\end{lemma}

We then show that for any atom in the original program (i.e., excluding the magic atoms), the formulas computed by $\Mcp$ only include correct choices wrt the original program. 
\begin{lemma}\label{lemma:claim5}
\ok{For 
an atom $a$}, let $\tau_a^i$ be the formula associated
with $a$ in $\Mcp^i(\emptyset)$. For every 
iteration $i$ and atom $a\in\Ap$ for which $\tau^i_a$ exists, for every total choice $C\subseteq\Fp$, 
 $C\models\tau_a^i \ok{\text{ implies that }} C\cup \Rp\models a$.
\end{lemma}

With these results at hand, we can prove the main statement. As $q\in\Ap$, by Lemma~\ref{lemma:claim5}, right-to-left holds for any $i$ for which $\tau^i_q$ exists, independently of $i_0$. Now consider a total choice $C$ such that $C\cup\Rp\models q$. By Theorem~\ref{thm:magic}, we have $C\cup\ok{\mathsf{magic}(\Rp,q)}\models q$. By Theorem~\ref{thm:ijar1} Point~2, we have that there is an $i_0$ such that for all $i\geq i_0$, $C\models\delta_q^i$. By Lemma~\ref{lemma:claim4}, we have that for all such $i$, $C\models\tau_q^i$. Thus, for every $C$, there is an $i_0$ such that left-to-right holds for any $i\geq i_0$, while at the same time right-to-left holds for any $i$, which completes the proof. QED.

\medskip \noindent
\textbf{Proof sketch of Lemma~\ref{lemma:claim4}.}  The proof is by induction on $i$. Note that both operators use the magic program. 

\textit{Base case.} In the first iteration, both variants assign formulas to 
facts in $\Fp$ only; for those, the formulas are identical. 

\textit{Inductive step.} Assume the claim holds for $i$, and consider $i+1$. For 
facts in $\Fp$, the same argument as for $i=1$ applies. For any magic atom $m$, as the claim holds for $i$, $\Tcm$ and $\Mcp$ apply the same rules to construct the formula for $m$, and as $\tau^{i+1}_m=\top$, the claim holds at $i+1$ for $m$. For any atom $a$ defined by a rule in $\Pp$,   as the claim holds for $i$, $\Tcm$ and $\Mcp$ apply the same rules to construct the formula for $a$, and for each of the body atoms in such a rule, the claim holds w.r.t. the formulas from iteration $i$, and thus also for the conjunctions of these formulas within a body, and the disjunctions across bodies for all active rules of $a$. QED.

\medskip \noindent
\textbf{Proof sketch of Lemma~\ref{lemma:claim5}.}    The proof is by induction on $i$. 
\textit{Base case.} The only $\tau^1$ that exist are those for 
facts, for which the claim is easily verified.
\textit{Inductive step.} Assume the claim holds for $i$, and consider $i+1$. For 
facts, the same argument as for $i=1$ applies.  Let $a\in\Ap$ be an atom defined by rules in $\Pp$ for which $\tau^{i+1}_a$ exists, and  $C$ a total choice for which $C\models\tau_a^{i+1}$.
By definition, $\tau^{i+1}_a = \bigvee_{(a,\varphi)\in B(\Ht^i,\Mp)}\varphi$, where ${(a,\varphi)\in B(\Ht^i,\Mp)}$ if and only if there is a rule $h\colonminus b_1,\ldots,b_n$ in $\Mp$ and a substitution $\theta$ such that $a=h\theta$ and for all $b_j$, there is a $\lambda_j$ with $(b_j\theta,\lambda_j)\in\Ht^i$, in which case $\varphi=\wedge_j\lambda_j$. 
As $C\models\tau_a^{i+1}$, there is at least one such rule with $C\models \wedge_j\lambda_j$. 
By definition of the magic program, for each such rule, $b_1$ is a magic atom, and thus $\lambda_1=\top$ and $C\models\lambda_1$, and $b_2,\ldots,b_n$ is the body of a rule for $h$ in $\Pp$, i.e., the $b_j$ are in $\Ap$. Thus, by assumption, if $C\models\wedge_j\lambda_j$ then for each $j>1$, $C\models b_j\theta$, and because $h\colonminus b_2,\ldots,b_n$ is in $\Rp$ and $a=h\theta$, $C\cup \Rp\models a$. QED.

\begin{figure*}
\begin{verbatim}
q01(X) :- graduatestudent(X), 
          takescourse(X,department0-university0-graduatecourse0) .
q02(X,Y,Z) :- graduatestudent(X), university(Y), department(Z), 
              memberof(X,Z), suborganizationof(Z,Y), 
              undergraduatedegreefrom(X,Y) .
q03(X) :- publication(X), 
          publicationauthor(X,department0-university0-assistantprofessor0) .
q04(X,Y1,Y2,Y3) :- professor(X), 
                   worksfor(X,department0-university0), 
                   name(X,Y1), emailaddress(X,Y2), telephone(X,Y3) .
q05(X) :- person(X), 
          memberof(X,department0-university0) .
q06(X) :-  student(X) .
q07(X,Y) :- student(X), course(Y), 
       takescourse(X,Y), 
       teacherof(department0-university0-associateprofessor0,Y) .
q08(X,Y,Z) :- student(X), department(Y), 
              memberof(X,Y), suborganizationof(Y,university0), 
              emailaddress(X,Z) .
q09(X,Y,Z) :- student(X), faculty(Y), course(Z), 
              advisor(X,Y), teacherof(Y,Z), takescourse(X,Z) .
q10(X) :- student(X), 
          takescourse(X,department0-university0-graduatecourse0) .
q11(X) :- researchgroup(X), 
          suborganizationof(X,university0) .
q12(X,Y) :- chair(X), department(Y), 
            worksfor(X,Y), suborganizationof(Y,university0) .
q13(X) :- person(X), 
          hasalumnus(university0,X) .
q14(X) :- undergraduatestudent(X) .
\end{verbatim}
\caption{LUBM queries.}
\label{fig:lubmqueries}
\end{figure*}

\end{document}